\documentclass[11pt]{article}

\usepackage{nodalida2021}
\usepackage{times}
\usepackage{hyperref}
\usepackage{latexsym}
\usepackage{graphicx}

\aclfinalcopy 

\title{Operationalizing a National Digital Library:\\ The Case for a Norwegian Transformer Model}

\author{Per E Kummervold \\ {\tt per.kummervold@nb.no}
    \And
    Javier de la Rosa \\ {\tt javier.rosa@nb.no}
    \AND
    Freddy Wetjen \\ {\tt freddy.wetjen@nb.no}
    \And
    Svein Arne Brygfjeld \\ {\tt svein.brygfjeld@nb.no}
    \AND
    \\ 
    The National Library of Norway \\ Mo i Rana, Norway
}
\date{}

\begin{document}
\maketitle
\begin{abstract}
In this work, we show the process of building a large-scale training set from digital and digitized collections at a national library. The resulting Bidirectional Encoder Representations from Transformers (BERT)-based language model for Norwegian outperforms multilingual BERT (mBERT) models in several token and sequence classification tasks for both Norwegian Bokm{\aa}l and Norwegian Nynorsk. Our model also improves the mBERT performance for other languages present in the corpus such as English, Swedish, and Danish. For languages not included in the corpus, the weights degrade moderately while keeping strong multilingual properties. Therefore, we show that building high-quality models within a memory institution using somewhat noisy optical character recognition (OCR) content is feasible, and we hope to pave the way for other memory institutions to follow.
\end{abstract}

\section{Introduction}

Modern natural language processing (NLP) models pose a challenge due to the massive size of the training data they require to perform well. For resource-rich languages such as Chinese, English, French, and Spanish, collections of texts from open sources such as Wikipedia \shortcite{wikiorg}, variations of Common Crawl data \shortcite{commoncrawl}, and other open-source corpora such as the BooksCorpus \citep{zhu15} are generally used. When researchers at Google released their Bidirectional Encoder Representations from Transformers (BERT) model, they trained it on a huge corpus of 16GB of uncompressed text (3,300M words) \citep{devlin18}. Later research has shown that the corpus size might have even been too small, and when Facebook released its Robustly Optimized BERT (RoBERTa), it showed a considerable gain in performance by increasing the corpus to 160GB \citep{liu19}.

Norwegian is spoken by just 5 million people worldwide. The reference publication \textit{Ethnologue} lists the 200 most commonly spoken native languages, and it places Norwegian as number 171. The Norwegian language has two different varieties, both equally recognized as written languages: Bokm{\aa}l and Nynorsk. The number of Wikipedia pages written in a certain language is often used to measure its prevalence, and in this regard, Norwegian Bokm{\aa}l ranges as number 23 and Nynorsk as number 55. However, there exist more than 100 times as many English Wikipedia pages as there are Norwegian Wikipedia pages \shortcite{metawiki}. When it comes to building large text corpora, Norwegian is considered a minor language, with scarce textual resources. So far, it has been hard to train well-performing transformer-based models for such languages.

As a governmental entity, the National Library of Norway (NLN) established in 2006 a mass digitization program for its collections. The Language Bank, an organizational unit within the NLN, provides text collections and curated corpora to the scholarly community \cite{sprakbanken}. Due to copyright restrictions, the publicly available Norwegian corpus consists mainly of Wikipedia pages and online newspapers, and it is around 5GB (818M words) in size (see Table \ref{tab1}). However, in this study, by adding multiple sources only accessible from the NLN, we were able to increase that size up to 109GB (18,438M words) of raw, deduplicated text. While such initiatives may produce textual data that can be used for the large-scale pre-training of transformer-based models, relying on text derived from optical character recognition (OCR)--based pipelines introduces new challenges related to the format, scale, and quality of the necessary data. On these grounds, this work describes the effort to build a pre-training corpus and to use it to train a BERT-based language model for Norwegian.

\subsection{Previous Work}

Before the advent of transformer-based models, non-contextual word and document embeddings were the most prominent technology used to approach general NLP tasks. In the Nordic region, the Language Technology Group at the University of Oslo, as part of the joint Nordic Language Processing Laboratory, collected a series of monolingual resources for many languages, with a special emphasis on Norwegian \citep{kutuzov17}. Based on these resources, they trained and released collections of dense vectors using word2vec and fastText (both with continuous skip-gram and continuous bag-of-words architectures) \citealt{mikolov13,bojanowski2017}, and even using an Embeddings from Language Models (ELMo)--based model with contextual capabilities \citep{peters18}. Shortly thereafter, Devlin et al. \shortcite{devlin18} introduced the foundational work on the monolingual English BERT model, which would later be extended to support 104 different languages including Norwegian Bokm{\aa}l and Norwegian Nynorsk, Swedish, and Danish. The main data source used was Wikipedia \shortcite{wikiorg}. In terms of Norwegian, this amounted to around 0.9GB of uncompressed text (140M words) for Bokm{\aa}l and 0.2GB (32M words) for Nynorsk \shortcite{metawiki}. While it is generally agreed that language models acquire better language capabilities by pre-training with multiple languages \cite{pires19,wudr20}, there is a strong indication that this amount of data might have been insufficient for the multilingual BERT (mBERT) model to learn high-quality representations of Norwegian at a level comparable to, for instance, monolingual English models \citep{pires19}.

In the area of monolingual models, the Danish company BotXO trained BERT-based models for a few of the Nordic languages using corpora of various sizes. Their repository \cite{botxo} lists models trained mainly on Common Crawl data for Norwegian (5GB), Danish (9.5GB), and Swedish (24.7GB). Unfortunately, we were unable to make the Norwegian models work, as they seem to be no longer updated. Similarly, the KBLab at the National Library of Sweden trained and released a BERT-based model and an A Lite BERT (ALBERT) model, both trained on approximately 20GB of raw text from a variety of sources such as books, news articles, government publications, Swedish Wikipedia, and internet forums \citep{malmsten20}. They claimed significantly better performance than both the mBERT and the Swedish model by BotXO for the tasks they evaluated.

At the same of the release of our model, the Language Technology Group at the University of Oslo released a monolingual BERT-based model for Norwegian named NorBERT. It was trained on around 5GB of data from Wikipedia and the Norsk aviskorpus \shortcite{nlpl}. We were unable to get sensible results when finetuning version 1.0 of their model. However, they released a second version shortly thereafter (1.1) fixing some errors \cite{norbert}. We have therefore included the evaluation results of this second version of the model in our benchmarking. They have also evaluated their and our model themselves \cite{KutBarVel21} with consistent results.

\section{Building a Colossal Norwegian Corpus}

As the main Norwegian memory institution, the NLN has the obligation to preserve and give access to all published information in Norway. A large amount of the traditional collection is now available in digital format. As part of the current legal deposit, many born-digital documents are also available as digital documents in the collection. The texts in the NLN collection span hundreds of years and exhibit varied uses of texts in society. All kinds of historical written materials can be found in the collections, although we found that the most relevant resources for building an appropriate corpus for NLP were books, magazines, journals, and newspapers (see Table \ref{tab1}). As a consequence, the resulting corpus reflects the variation in the use of the Norwegian written language, both historically and socially. 

\begin{figure*}
\centering
\includegraphics[width=0.9\textwidth]{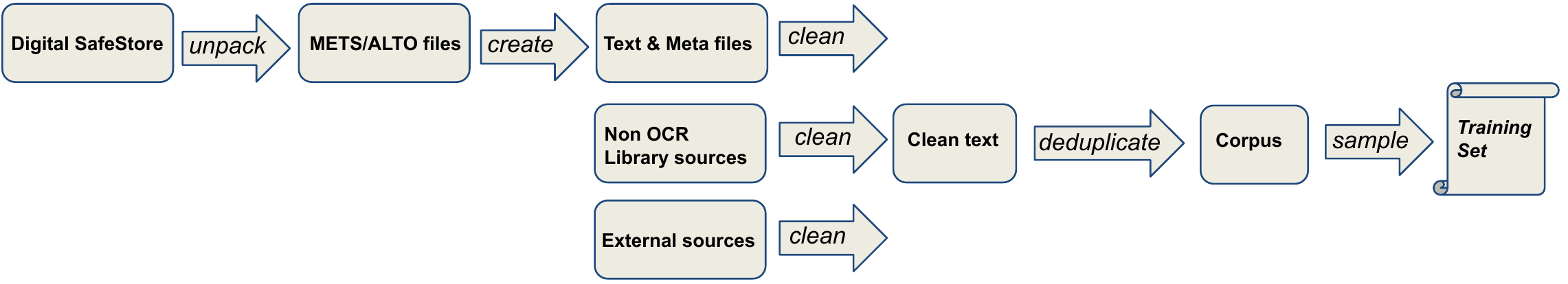}
\caption{The general corpus-building process.}\label{fig1}
\end{figure*}

Texts in the NLN have been subject to a large digitization operation in which digital copies were created for long-term preservation. The NLN employs METS/ALTO\footnote{Metadata Encoding and Transmission Schema and Analyzed Layout and Text Object \cite{mets,alto}} as the preferred format for storing digital copies. As the digitized part of the collection conforms to standard preservation library practices, the format in which the texts are stored is not suitable for direct text processing; thus, they needed to be pre-processed and manipulated for use as plain text. One major challenge was the variation in the OCR quality, which varied both over time and between the types of materials digitized. This limited the number of usable resources and introduced some artifacts that affected the correctness of the textual data.

The basic inclusion criterion for our corpus was that as long as it was possible for a human to infer the meaning from the text, it should be included. However, the amount of text involved in building the model meant that this needed to be determined automatically. The METS/ALTO files contain information from the OCR process regarding the confidence of every word (from 0 for no confidence to 1 for certainty), so we used this assessment to calculate the average confidence for paragraphs and pages. Setting the minimum paragraph confidence to 0.8 and the minimum page confidence to 0.9 allowed us to filter out a significant part of the text with the lowest quality. We also noticed that in the period of digitization from the beginning of 2006 until the end of 2008, the quality of the OCR was low and the estimated confidence values were too optimistic. We ended up excluding all text scanned in this period. 

To further filter out erroneous textual information, we calculated the number of words in the documents and averaged the number of words per paragraph. Establishing a threshold of at least 20 words per document and an average of 6 words per paragraph, we could filter out text sources that had little value for training, such as cartoons and picture books. We estimated the language composition using various methods, including metadata tags in the collection and counting the frequency of words of certain types (e.g., personal pronouns). Our estimate is that 83$\%$ of the text is in Norwegian Bokm{\aa}l and 12$\%$ is in Nynorsk. Close to 4$\%$ of the texts are written in English, and the 1$\%$ left is a mixture of Sami, Danish, Swedish, and a few traces from other languages. 

The aforementioned process was carefully orchestrated, with data moving from preservation storage, through error correction and quality assessment, and ending up as text in the corpus. As shown in Figure \ref{fig1}, after filtering, OCR-scanned documents were added to the other digital sources. After this step, the data went through the cleaning process, in which we ensured the consistency of the text encoding and special characters used. In the deduplication stage, all duplicated paragraphs in the entire collection were removed. Finally, we drew out two pre-training-sets: one with a sequence length of 128 tokens, and one with a sequence length of 512 tokens.

\begin{table*}[ht]
\begin{center}
\begin{tabular}{llrr}
\hline
\textbf{Sources}&\textbf{Period}&\textbf{Words (Millions)}&\textbf{Text (GB)}\\
\hline
Books (OCR) & 1814--2020 & 11,820 & 69.0\\
Newspaper Scans (OCR) & 2015--2020 &3,350 & 20.0\\
Parliament Documents$^{a}$ (OCR) & 1814--2014 & 809 &5.1\\
Common Crawl OSCAR & 1991--2020 & 799 & 4.9\\
Online Bokm{\aa}l Newspapers & 1998--2019 & 678 &4.0\\
Periodicals (OCR) & 2010--2020&317&1.9\\
Newspaper Microfilms (OCR) & 1961, 1971, 1981, 1998--2007 & 292 & 1.8\\
Bokm{\aa}l Wikipedia & 2001--2019 & 140 &0.9\\
Public Reports$^{b}$ (OCR) & 1814--2020 &91&0.6\\
Legal Collections$^{c}$ & 1814--2004 &63&0.4\\
Online Nynorsk Newspapers & 1998--2019 & 47 & 0.3\\
Nynorsk Wikipedia & 2001--2019 & 32 & 0.2\\
\hline
Total (After Deduplication) && 18,438 & 109.1\\
\hline
\multicolumn{4}{l}{
$^{a}$\footnotesize{Stortingsforhandlingene.}
$^{b}$\footnotesize{Evalueringsrapporter.}
$^{c}$\footnotesize{Lovdata CD/DVD.}
} \\
\end{tabular}
\end{center}
\caption{\label{tab1} The composition of the Colossal Norwegian Corpus.}
\end{table*}

\section{Pre-training a Norwegian BERT model}

In order to build our own pre-trained language model for Norwegian, we decided to use the original BERT architecture pre-trained with a masked-language model (MLM) objective, as published by Devlin et al. \shortcite{devlin18}. We evaluated the effect of changes in hyperparameters in terms of MLM performance and of the fine-tuning of the pre-trained models on various downstream tasks. All pre-training work was run on a v3-8 TPU (128GB) provided by the TPU Research Cloud, while the evaluation was done on in-house machines with a single NVIDIA Quadro RTX6000 (24GB).

Our goal was to build a solid model that would perform well on all types of Norwegian language tasks, ranging from old to modern text, and including texts that might be mixed with foreign languages like English. We therefore chose to initiate the model from the pre-trained mBERT weights \cite{tfhub}. The mBERT model was trained on 104 languages, including both Norwegian varieties (Bokm{\aa}l and Nynorsk). The model uses a 119,547-token vocabulary, and its pre-trained weights might also benefit from cross-lingual transfer. Our assumption is that using the mBERT weights for Norwegian should result in a better-performing model in comparison to starting with random weights. It might also keep some of its multilingual abilities, making it more robust when dealing with new words and texts containing fragments of other languages \citep{wudr20}.

\subsection{Improving the Model Beyond mBERT}

\begin{figure}
\centering
\includegraphics[scale=.18]{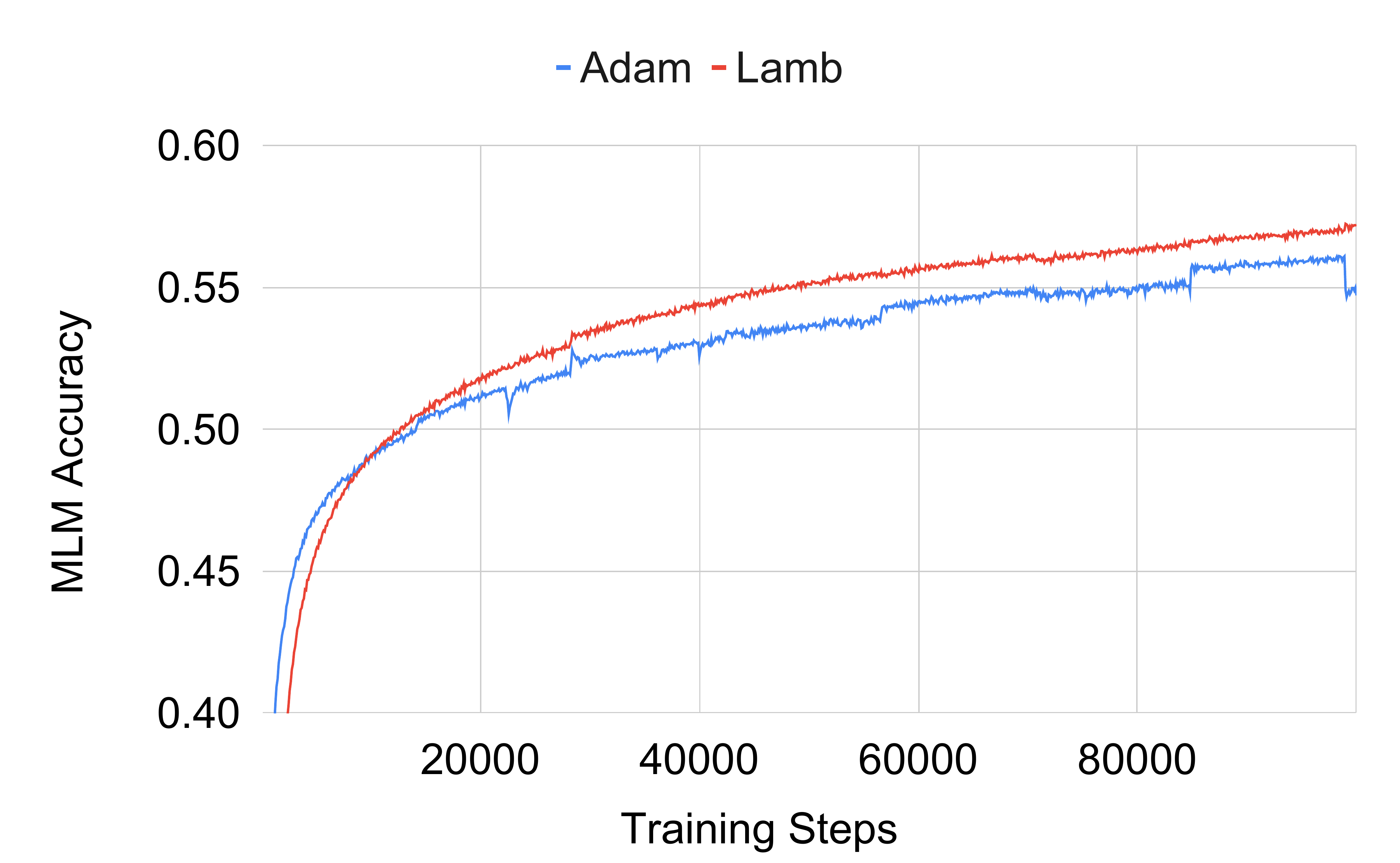}
\caption{Comparison of Adam and LAMB optimizers (learning rate: 4e-4; batch size: 2,048).}\label{fig2}
\end{figure}

All subsequent training runs followed the findings by You et al. \shortcite{ylrh19}, who showed that the pre-training of a BERT model could be improved by increasing the batch size but that, at the same time, an increase in the learning rate could lead to instability, especially when using the adaptive moment estimation (Adam) optimizer. When training on large batch sizes, You et al. suggested using their layer-wise adaptive moments base (LAMB) optimizer instead. We confirmed these results on our dataset when pre-training for 100,000 steps on a batch size of 2,048 sequences, which is very close to the optimum size for our v3-8 TPU (128GB) setup (see Figure \ref{fig2}).

The basic pre-training strategy was to use the largest possible batch size on our TPU and to increase the learning rate as long as it showed stability. An evaluation of the learning rate was done for 100,000 steps, but because we used decay, we expected the stability to be maintained even after this point. Devlin et al. \shortcite{devlin18} trained for 128-length sequences for approximately 90$\%$ of the training examples, then trained for 512-length sequences for 10$\%$ . Due to memory limits on our TPUs, we needed to reduce the batch size (by a factor of approximately 7) for the 512 sequences in the pre-training data; we also increased the number of pre-training steps for the long sequences to resemble the same distribution of short and long sequences that were used in training the BERT model. To investigate the effect of this, we experimented with two different setups in our model (version A and version B). Both were initialized from the same mBERT weights and trained identically for the first 1,750,000 steps. In the last steps, version A followed the training schedule used in the BERT model where roughly 10$\%$ of the total training time was used on long sequences (step 3a) and then an additional step (3b) on shorter sequences. Version B reduced the training on short sequences and instead trained almost 30$\%$ of the time on long sequences. The setup was chosen for making the total training time roughly the same for both models (see Table \ref{tab2}). 

\begin{table*}[ht]
\begin{center}
\resizebox{\textwidth}{!}{
\begin{tabular}{lrrrrrr}
\hline
\multicolumn{4}{l}{}& \multicolumn{2}{c}{\textbf{Version A}} & {\textbf{Version B}} \\
& {\textbf{Warmup}} & {\textbf{Step 1}} &{\textbf{Step 2 }} &
{\textbf{Step 3a}} &{\textbf{Step 3b}} & {\textbf{Step 3}} \\
\hline
{Steps} & {50k} & {700k} &{1M} &{1.2M} &{1.2M} &{2M} \\
{Batch Size} &{2760} &{2760} &{384} &{384} &{2760} &{384} \\
{Examples} &{138M} &{1,938M} &{384M} &{460M} &{3,312M} &{768M}\\
{Sequence Length} &{128} &{128} &{512} &{512} &{128} &{512} \\
{Learning Rate} &{0 $ \rightarrow $ 4e-4 } &{4e-4} &{4e-4} &{ 4e-4 $ \rightarrow $ 2e-4 } & {2e-4 $ \rightarrow $ 0} & {4e-4 $ \rightarrow $ 0} \\
\hline
\end{tabular}}
\end{center}
\caption{\label{tab2} Training schedule for our models. }
\end{table*}

\section{Evaluation}

While pre-trained language models also can be used for direct MLM-predition and feature extractions, the most common use is to fine-tune it on a specific task. The base procedure for fine-tuning was described by \citet{vaswani17}, and it consists of training for a small number of epochs (typically 4), with a warmup of around 10$\%$ of the training steps; subsequently, a linear decay to zero is used. Devlin et al. \shortcite{devlin18} based their work on the same procedure and selected the best learning rate among 5e-5, 3e-5, and 2e-5, according to the performance of the model on the validation set. The optimal learning rate and number of epochs mainly depend on the size of and variance in the training corpus, but they can also be affected by the properties of the pre-trained model. To get optimal performance out of a pre-trained model, the hyperparameters in the fine-tuning should be adapted. However, in this work, we are not primarily interested in optimization but in a comparison of the performance of our models against the mBERT model.

\subsection{Token Classification}

A common way to evaluate language models is by fine-tuning the models on token classification tasks such as named-entity recognition (NER) and part-of-speech (POS) tagging. For Norwegian, the Norwegian Dependency Treebank \citep[{NDT,}][]{norwegian21} by the Spr{\aa}kbanken at the NLN and the Language Technology Group at the University of Oslo provide text that has been manually annotated with morphological features, syntactic functions, and hierarchical structures. The morphological annotation mainly follows the Oslo-Bergen tagger \citep{johannessen12}, and with a few exceptions, the syntactic analysis follows the Norwegian Reference Grammar \citep{faarlund1997norsk}.\ With the help of Schibsted Media Group, the same group recently published Norwegian Named Entities (NorNE) \cite{JorAasHus20}, an extension of NDT that includes named-entity annotations for more than 300,000 tokens. 

Moreover, with the goal of testing being the retaining or vanishing of the multilingual abilities of our model, we also considered NER datasets in both languages included in our corpus and in languages of which there is little to no evidence in our corpus. Specifically, we used CoNLL-2003 for English \citep{tjong-kim-sang-2003-introduction}, Webbnyheter 2012 for Swedish \citep{webbnyheter2012}, DaNE for Danish \citep{hvingelby20}, CoNLL-2002 for Spanish \citep{tjong-kim-sang-2002-introduction}, and FiNER for Finnish \citep{ruokolainen19}. While the number and specificity of the tag sets vary across datasets, rendering the comparison between languages useless, we could still compare the performance of our model against that of English-only and multilingual BERT models. We decided to leave out NER datasets built using automated or semi-automated annotations processes.

\subsection{Sequence Classification}

For sequence classification, we chose another commonly used task: sentiment classification. We used a version of the Norwegian Review Corpus (NoReC) \citep{ovrelid-etal-2020-fine}, a fine-grained sentiment dataset \citep{nosent} for Norwegian created by the Nordic Language Processing Laboratory. Moreover, we defined a second sequence-classification task to capture idiosyncrasies and nuances of the Norwegian language. In this case, we generated a balanced corpus of 6,000 text speeches that had been spoken at the Norwegian Parliament (Storting) between 1998 and 2016 by members of the two major parties, Fremskrittspartiet and Sosialistisk Venstreparti \cite{LapSoyVel18}. The dataset is annotated with the party the speaker was associated with at the time, and the source data was made publicly available by the Norwegian parliament. The classification task is to determine the political affiliation of the transcribed speech segment. 

\section{Results}

To evaluate the performance of our model, we searched for the optimal set of fine-tuning hyperparameters for each downstream task by running a small grid search (see Table \ref{tab3}) on the mBERT model. The search space was the same for all tasks and included learning rates ranging from 2e-5 to 5e-5, with the number of training epochs being 3 or 4. We did the same for the warmup ratio and weight decay. The performance was generally best using a warmup ratio of 0.1 and weight decay of 0, so we applied this universally to limit the grid complexity.

For the token classification tasks, we selected the best-performing hyperparameters based on the seqeval \shortcite{seqeval18} F1 micro score on the validation set for Bokm{\aa}l after fine-tuning an mBERT model. For sequence classification, we used the F1 macro score.

\begin{table}[ht]
\begin{center}
\resizebox{\columnwidth}{!}{
\begin{tabular}{lrrrr}
\hline
&{\textbf{NER}} & {\textbf{POS}} & {\textbf{Sentiment}} &{\textbf{Political}} \\
\hline
{Learning Rate} & 2e-5 & 3e-5 & 3e-5 & 2e-5 \\
{Number of Epochs} & 3 & {3} & {3} & {3} \\
\hline
\end{tabular}}
\end{center}
\caption{\label{tab3} Optimal fine-tuning hyperparameters for the mBERT model using the validation datasets.}
\end{table}

We then used the optimal fine-tuning parameters from the mBERT model for the validation dataset on our model and on the NorBERT model. Last, we compared all the models based on their results in relation to the test dataset.

Version B of our model---the version with the extended training-sequence length---performed slightly better on all four tasks than did version A. To simplify the results presented here, we therefore report only the results from version B, which we are naming \textit{NB-BERT}.

\begin{table*}[ht]
\begin{center}
\begin{tabular}{lrrrrrr}
\hline
& \multicolumn{2}{c}{NER} & \multicolumn{2}{c}{POS} & {Sentiment} & {Political}\\

\hline
& {\textbf{Bokm{\aa}l}} & {\textbf{Nynorsk}} & {\textbf{Bokm{\aa}l}} & {\textbf{Nynorsk}} & {\textbf{Bokm{\aa}l $\&$ Nynorsk}} & {\textbf{Bokm{\aa}l}} \\
\hline
{mBERT} & {83.8} & {85.6} & {98.3} & {98.0} & {69.7} & {78.4} \\
{NorBERT} & {89.9} & {86.1} & {98.5} & {98.4} & {81.7} & {78.2} \\
{NB-BERT (ours)} & {\textbf{91.2}} & {\textbf{88.9}} & {\textbf{98.8}} & {\textbf{98.8}} & {\textbf{86.4}} & {\textbf{81.8}} \\
\hline
\end{tabular}
\end{center}
\caption{\label{tab4} Evaluation results from the test dataset (version B of the model; F1 micro in token classifications and F1 macro in sequence classifications; best scores in bold).}
\end{table*}

As can be seen in the Table \ref{tab4}, the NB-BERT model performed significantly better than did the mBERT model for both Bokm{\aa}l and Nynorsk, and on both token and sequence classification. The improvement was the smallest for the POS dataset, with an improvement from 98.3 to 98.8 for Bokm{\aa}l and from 98.0 to 98.8 for Nynorsk. However, POS datasets such as this always contain some ambiguity, and it is hard to tell how much more improvement is possible there. In addition, the NER task improved from 83.8 to 91.2 for Bokm{\aa}l and from 85.6 to 88.9 for Nynorsk. The sequence classification improved from 69.7 to 86.4 in terms of sentiment classification and from 78.4 to 81.8 for political classification. We also tested the release 1.1 of the NorBERT model that is uploaded to Hugging Face \citep{norbert}. The performance of this model lays in between that of NB-BERT and mBERT for Bokm{\aa}l and Nynorsk, but it generally performs worse on all non-Norwegian tasks.\par

As shown in Table \ref{tab5}, our model was able to outperform the English-only and multilingual BERT for both Norwegian Bokm{\aa}l and Nynorsk, as well as for Swedish and Danish, which are languages with a shared tradition with Norwegian. For English, our results are also marginally better than those obtained using the English-only BERT model. For Spanish and Finnish, for which there is no close relationship with Norwegian nor documented occurrences of text in such languages in our corpus, the mBERT model outperformed both the English-only BERT and our model, suggesting that our model is deteriorating for the languages not included in the corpus.\par

\begin{table*}[ht]
\begin{center}
\begin{tabular}{lrrrrrrr}
\hline
&{\textbf{Bokm{\aa}l}} &{\textbf{Nynorsk}} & {\textbf{English}} & {\textbf{Swedish}} & {\textbf{Danish}} & {\textbf{Spanish}} & {\textbf{Finnish}} \\
\hline
{English BERT} & {75.1} & {77.8} & {91.3} & {82.5} & {73.9} & {81.8} & {82.9} \\
{mBERT} & {83.8} & {85.6} & {90.8} & {85.3} & {83.4} & {\textbf{87.6}} & {\textbf{88.7}} \\
{NorBERT} & {89.9} & {86.1} & {87.8} & {83.4} & {80.7} & {79.3} & {81.5} \\
{NB-BERT (ours)} & {\textbf{91.2}} & {\textbf{88.9}} & {\textbf{91.3}} & {\textbf{85.9}} & {\textbf{85.1}} & {85.8} & {85.8} \\
\hline
\end{tabular}
\end{center}
\caption{\label{tab5} Evaluation results (F1 micro) of different monolingual NER datasets using the English-only BERT, mBERT, NorBERT, and our model (best scores in bold).}
\end{table*}

\section{Discussion}

The majority of the training corpora used today for training transformer models are built using mainly open web sources. A major motivation for this project was to investigate whether the digital collections at the NLN could be used to create a suitable corpus to train state-of-the-art transformer language models. The texts available through the library are heterogeneous in nature, including cartoons, novels, news articles, poetry, and government documents published over time and in different contexts. As our results suggest, this seems to be a strength rather than a weakness, in that it enables us to build high-performance transformer models for small languages, such as Norwegian. Consequently, our Norwegian corpus is not only richer in diversity but also significantly larger in size than is any other Norwegian corpus, and it even rivals the size of previous work on a major language such as English. The Norwegian part of the mBERT model consists of around 1GB of text \citep{wudr20}, while the English-only BERT model was trained on 16GB of text \citep{devlin18} mainly based on English Wikipedia and Open Book Corpus. When Facebook developed the first version of its RoBERTa, it added Common Crawl data and Open WebText to the BERT corpus and ended up with 160GB of text \citep{liu19}. Our clean corpus of Norwegian-only text is 109GB in size.

For the target languages Norwegian Bokm{\aa}l and Norwegian Nynorsk, the model performs significantly better than does the mBERT model on both token classifications (POS and NER) as well as on the two sequence classification tasks. In the Bokm{\aa}l NER task, the level of improvement was +7.4 F1 points. Because none of the datasets have been benchmarked against human performance, it is hard to measure how close this is to the theoretical maximum.

The results show that our corpus is a valid training source, and this is by no means surprising. All research points to the possibility of improving transformer models' performance by training them on larger text corpora. However, the novelty of our results lies in that we were able to increase the performance on our domain-specific tasks while maintaining a lot of the multilingual properties of the mBERT model. This was unexpected because English only comprised around 4$\%$ of the training set. Still, we were able to improve the English capabilities of the model up to the level of the monolingual English model. Part of the reason for this might be that we applied some training techniques that were not available when the English-only model was trained and released, most notably the use of larger batch sizes and the LAMB optimizer.

We were also able to significantly improve the scores for Swedish and Danish, though it is hard to pinpoint how much of this was caused by the close linguistic similarities between the languages and how much by the fact that they were represented in the corpus to some degree.

It should not be surprising that the capabilities of the model in relation to languages that were not included in the training corpus (i.e., Spanish and Finnish) did deteriorate. However, the drop in performance was not radical, and the results above indicate that we might have been able to prevent this by adding just a small portion of these languages to the large corpus.

Overall, our results suggest that collections such as the digital collection at the NLN, even if they contain occational OCR-errors, may contribute significantly toward the creation of well-performing language models by providing large training corpora. As discussed earlier, there are OCR errors in the included materials. An exhaustive removal of all OCR artifacts would either have required us to do a major reduction of the size of the corpus, or to invest an unmanageable amount of manual work. We have not seen any indication that the OCR errors negatively impacted the performance. We might speculate that the model has learned to distinguish OCR errors from ordinary text, indicating that quantity is more important than quality when building such corpora. All in all, size matters.

\section{Conclusion and Future Work}

In this work, we have investigated the feasibility of building a large Norwegian-only corpus for the training of well-performing transformer-based language models. We relied on the collections of the NLN, and our model outperformed the existing multilingual alternatives. In the process, while the corpus produced might lack the cleanness of other textual resources, we proved that using somewhat noisy but available sources is an effective way to grow the ecosystem of resources for languages with fewer resources and for which enough open text in a digital format simply does not exist. As part of an effort to democratize the use of technology and digital resources at the NLN, we are releasing our trained BERT-based model \citep{NbAiLab21a} and will be releasing other models based on the same corpus in the future. Moreover, we are also releasing the set of tools and code we used so that others seeking similar results can easily reuse them \citep{NBAiLab21b}.

Although our work may indicate that OCR errors in corpora have little to no impact on the quality of the resulting transformer model, this has not been explicitly proven in the current study. More systematic studies are needed to investigate the real effect of OCR noise and artifacts. 

Another important aspect is that, to benefit from the pre-trained mBERT weights, we used a 119,547-token multilingual vocabulary, of which only a small fraction pertained to Norwegian. A natural follow up would be to investigate the performance gains of using only a tailored Norwegian vocabulary.

The decision to use a BERT-based architecture as our target was guided by its simplicity to train and benchmark. However, newer and better-performing models have been released since the original BERT work a few years ago. The current corpus could be used for training such models as well studying the differences between architectural styles and training objectives. While it is already large in size, there is still potential to grow our 109GB corpus to the limits of the extant Norwegian holdings at the NLN, which presents itself as an opportunity to release even larger models.

\section*{Funding}

This research was supported by Cloud TPUs from Google's TPU Research Cloud (TRC).

\section*{Acknowledgment}

We would like to thank KBLab at the National Library of Sweden (Kungliga biblioteket) for its pioneering work on BERT in memory institutions and for the valuable and inspiring discussions. We also appreciate the feedback from and discussions with Andre Kaasen of the Language Bank at the National Library of Norway.

\bibliographystyle{acl_natbib}
\bibliography{norwegian_transformer}

\end{document}